\documentclass{article}
\usepackage{spconf,amsmath,graphicx}
\usepackage{float}
\usepackage[table,xcdraw]{xcolor}
\usepackage{enumitem}
\usepackage{lipsum}

\setlist{labelsep=*} %% set here.
\newcommand\blfootnote[1]{%
  \begingroup
  \renewcommand\thefootnote{}\footnote{#1}%
  \addtocounter{footnote}{-1}%
  \endgroup
}

\title{Contextually-rich human affect perception using multimodal scene information} 
\name{Digbalay Bose, Rajat Hebbar, Krishna Somandepalli $^{\dagger}$, Shrikanth Narayanan}
\address{Signal Analysis and Interpretation Laboratory, University of Southern California, CA 90089 \\
\tt\small {dbose@usc.edu,rajatheb@usc.edu,somandep@usc.edu,shri@ee.usc.edu}}

\begin{document}
\ninept
\maketitle
\def\thefootnote{$\dagger$}\footnotetext{The work was done while the author was at USC}\def\thefootnote{\arabic{footnote}}

\begin{abstract}
The process of human affect understanding involves the ability to infer person specific emotional states from various sources including images, speech, and language. Affect perception from images has predominantly focused on expressions extracted from salient face crops. However, emotions perceived by humans rely on multiple contextual cues including social settings, foreground interactions, and ambient visual scenes. In this work, we leverage pretrained vision-language (VLN) models to extract descriptions of foreground context from images. Further, we propose a multimodal context fusion (MCF) module to combine foreground cues with the visual scene and person-based contextual information for emotion prediction. We show the effectiveness of our proposed modular design on two datasets associated with natural scenes and TV shows.
\end{abstract}
\section{Introduction}
\label{sec:intro}
There has been increased interest in understanding the affective processes associated with various facets of human emotions \cite{dukes2021}. An integral part of affective understanding is the ability to infer expressions of human emotions from various sources like images \cite{AICA}, speech \cite{speechemo} and language use. Affect recognition systems have enabled multiple human-centered applications, notably in healthcare (depression detection \cite{depressiondetection}, autism-spectrum diagnosis \cite{autismguha}) and learning \cite{savchecnkoengagement}. 
\par
Affect recognition from images has largely focused on facial expressions \cite{DFEW,Mollahosseini2019AffectNetAD} along a fixed set of categories. Moreover, facial expression based methods typically consider crops of a single face, which might provide ambiguous signals for classifying perceived emotions. Emotion perception in humans typically relies on multimodal behavioral cues, that go beyond facial expressions, such as voice and language \cite{busso2004analysis}. However, are there additional contextual cues beyond behavioral expressions, such as of face and language, that mediate human emotion perception? Studies have shown that contextual information including the social setting, interaction type, and ambient location can play a key role \cite{BarretEmotionPerception}. Context in images is driven by visual scenes \cite{Bar2004VisualOI} or specific locations such as outdoor, indoor,  kitchen, living room etc and the interactions between the various entities in the scene.
\begin{figure}[h!]
    \centering
    \includegraphics[width=\columnwidth]{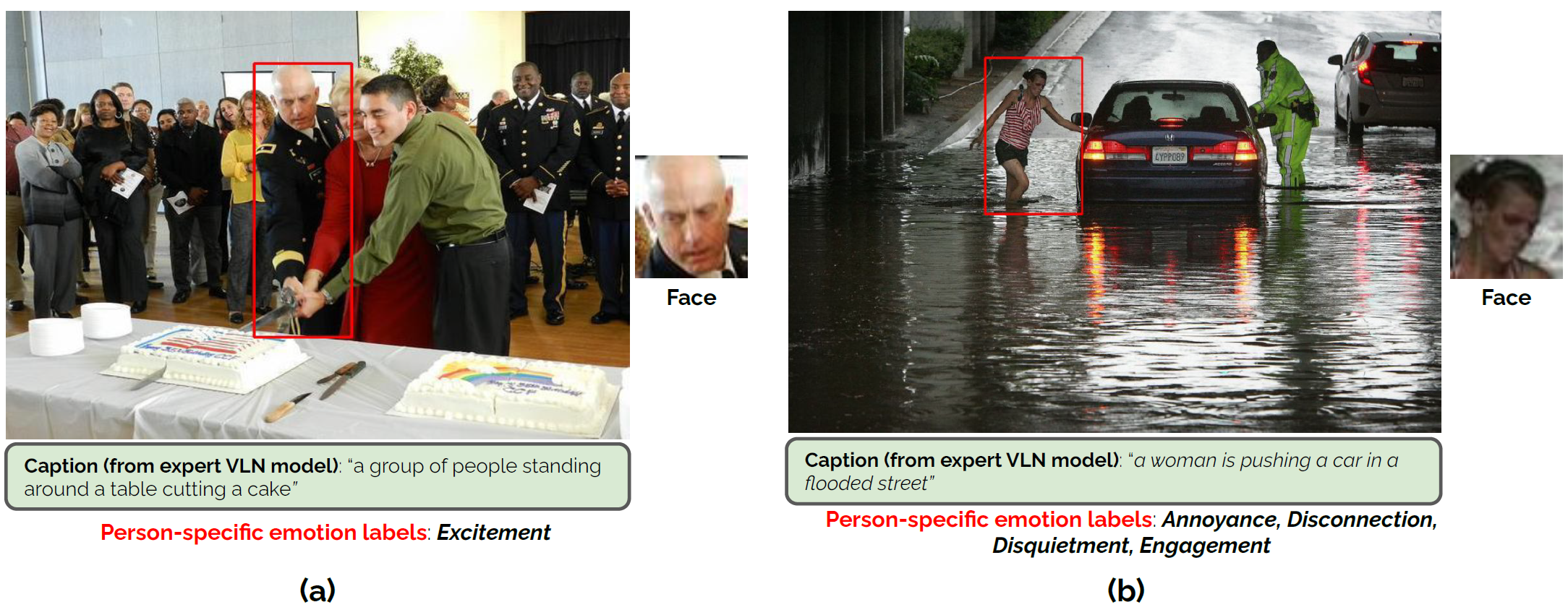}
    \caption{Examples from EMOTIC dataset showing the importance of context for estimating person specific emotion labels. Extracted captions from a pretrained VLN model (OFA) capture the foreground contexts.}
    \label{intro_context}
\end{figure}
An example is shown in Fig.\ref{intro_context} with respect to both (a) positive and (b) negative emotions. In Fig.\ref{intro_context} (a), the face crop provides a negative signal whereas the overall scene including the generated caption from a pretrained vision-language model OFA \cite{wang2022ofa} indicates a positive event associated with the person. In Fig. \ref{intro_context}(b), while the face crop provides a noisy, incomplete signal for perceiving the expressed emotional state, the overall context of the visual scene plus its descriptive caption indicates the distressing situation associated with street flooding. 

\par 
Recent advances in multimodal vision-language (VLN) pretraining \cite{cho2021vlt5, wang2022ofa} has resulted in task-agnostic transformer-based models that can be used for variety of tasks such as image captioning and visual question answering. In this work, we employ the pretrained VLN models as experts for describing foreground context in terms of captions. Further we consider contextual information in terms of the individual persons (whole-body appearance or face) and the visual scenes. In order to effectively leverage multiple contextual sources we propose attention-based multimodal context fusion (MCF) module to predict both discrete emotion labels and continuous-valued arousal, valence and dominance values. We show the effectiveness of our methods on two publicly available datasets: EMOTIC \cite{kostiPAMI} (natural scenes) and CAER-S \cite{CAER-S}(TV shows).
\blfootnote{Code:https://github.com/usc-sail/mica-context-emotion-recognition}
\section{Related work}
\label{sec:format}
\textbf{Context in Emotion Perception:} The role of context extraneous to a person (beyond their traits and behavior) in the perception of their expressed emotion has been studied from the perspective of scenes \cite{BarretEmotionPerception}, and cultures \cite{Masuda2008PlacingTF}. In \cite{Dudzik2019ContextIH}, the perceivable-encoding context and the prior knowledge available with the perceivers are reported as the major sources of context for influencing emotion perception.  
Situational context like reactions from other people has been considered in \cite{Wieser2012FacesIC} as a means to decode emotional states of persons in consideration.\\
\textbf{Context-based image datasets:} Image-based emotion recognition datasets like AffectNet \cite{AffectNet}, FER \cite{BarsoumICMI2016}, DFEW \cite{DFEW} primarily focus on signals encoded in facial expressions. Since emotion perception depends on where the facial configuration is present, datasets like EMOTIC \cite{kostiPAMI} and CAER \cite{CAER-S} have been proposed to incorporate contextual information in terms of visual scenes and social interactions. In the case of EMOTIC, annotators have marked person instances in unconstrained environments with apparent emotional states based on the existing scene context. However, the annotation process in CAER revolves around TV shows with primary focus on interactions-driven context. 
\\
\textbf{Context modeling approaches:} 
\cite{kostiPAMI,CAER-S} explore context modeling in terms of dual stream models for processing body and whole image streams. \cite{CAGER} also uses a dual stream network with context stream, modeled using an affective graph composed of region proposals.\cite{Mittal2020EmotiConCM} uses depth and pose as additional contextual signals with existing streams like scene, face for predicting person specific emotion .
\cite{pikoulis2021leveraging} explores contextual modeling in short movie clips by considering scene and action characteristics along with body (including face) based signals. In contrast, our approach uses natural language descriptions to describe the foreground context along with scene and person specific streams.
\\
\textbf{Multimodal vision-language models:} Vision language (VLN) models like OFA \cite{wang2022ofa}, VL-T5 \cite{cho2021vlt5}, ALBEF \cite{ALBEF} are pretrained on large-scale image-text pairs curated from the web, thus enabling its usage in diverse tasks like image captioning, retrieval, visual-question answering etc. In our formulation, we harness the capabilities of VLN models to generate descriptive captions since they contain condensed description of the foreground context in terms of entities including persons.
\section{Problem formulation}
 Given an image $I$ and a person-specific context in the form of bounding box $[x,y,w,h]$, the task is to predict the emotional state $p$ associated with a person as $p_{disc}, p_{cont} = F(I,[x,y,w,h])$. $p_{disc}$ and $p_{cont}$ refer to the predicted set of discrete emotion categories and continuous arousal, valence and dominance values, respectively. $F$ refers to the deep neural network used for estimating the discrete and continuous affective states. The design of $F$ is dependent on extraction of multiple contextual information from the given image $I$, that are listed as below:\\
 \textbf{\underline{Visual scene context}}: The underlying visual scene ($VS$) (e.g.,  kitchen, bar, football field etc) plays a role in influencing the emotional state of a person. Here we use a ViT \cite{Dosovitskiy2021AnII} model ($f_{VS}$) finetuned on Places365 \cite{zhou2017places} as the backbone network for extracting visual scene representations ($e_{VS}$) from $I$. \\
 \textbf{\underline{Person context}}: The person-specific context is extracted using a whole-body or facial bounding box, denoted by $[x,y,w,h]$ from image $I$. The cropped person instance is passed through a person encoder ($f_{PE}$) i.e. Resnet34~\cite{He2016DeepRL} for extracting person-centric representations ($e_{PE}$). \\
 \textbf{\underline{Language driven foreground context}}: Natural language description of image $I$ provides foreground (FG) context in terms of entities including persons and their interactions. We use a 12-layer transformer encoder-decoder model $OFA_{large}$ \cite{wang2022ofa} as $f_{expert}$ to extract the foreground specific captions for image $I$. For extracting text representations ($e_{FG}$) of the captions, we use BERT's \cite{Devlin2019BERT} pretrained encoder ($f_{FG}$) from HuggingFace \cite{wolf-etal-2020-transformers}.
\section{Multimodal Context Fusion (MCF) Module}
\begin{figure}
    \centering
    \includegraphics[width=0.9\columnwidth]{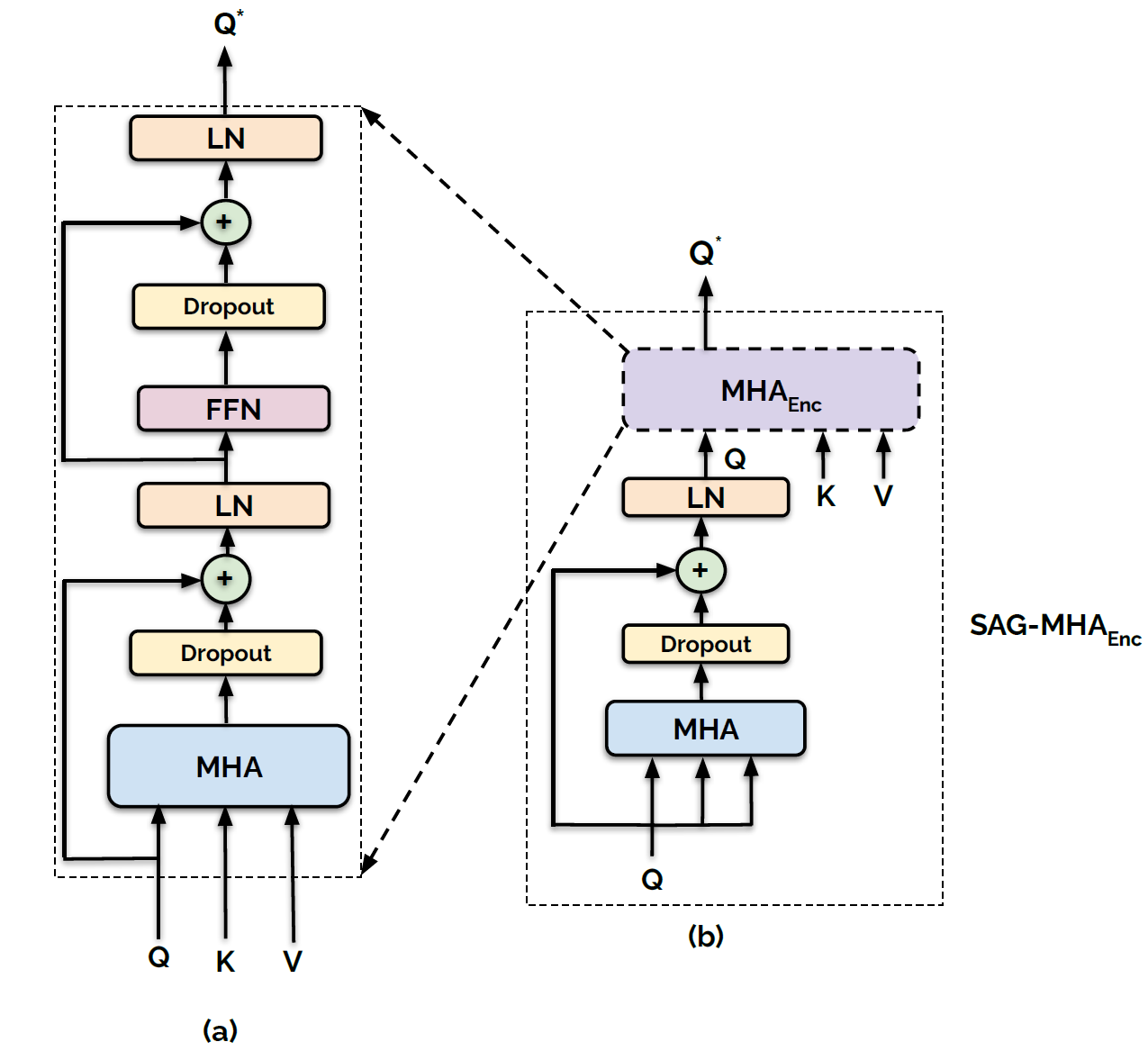}
    \caption{Outline of (a) MHA\textsubscript{enc} (b) SAG-MHA\textsubscript{enc} layers. \textbf{LN}: Layer Norm, \textbf{FFN}: Feedforward network, \textbf{MHA:} Multihead attention, \textbf{SAG:} Self Attention Augmented}
    \label{encoder block}
\end{figure}
The multimodal context fusion module is composed of two parallel streams, associated with foreground and visual scene based contexts. The basic operation in individual streams is a cross-modal encoder block CM\textsubscript{enc} composed of L encoder layers. As shown in Fig \ref{encoder block}, we consider two designs for the encoder layer i.e., MHA\textsubscript{enc} and SAG-MHA\textsubscript{enc}.  
The set of operations in encoder MHA\textsubscript{enc} layer for query ($Q$), key ($K$) and value ($V$) representations are listed as follows:
\begin{equation} \label{MHAEnc operation}
\begin{split}
 Q^{'} & = LN(Q + \text{Dropout}(MHA(Q,K,V))) \\ 
Q^{*} & =LN(\text{Dropout}(FFN(Q^{'}))+Q^{'})
\end{split}
\end{equation}
Here $MHA$, $LN$ and $FFN$ refer to Multi-head attention, layer-norm operation and feed-forward neural network respectively. The SAG-MHA\textsubscript{enc} layer consists of a multi-head attention based transformation of the query representations followed by input to the MHA\textsubscript{enc} layer. The design of SAG-MHA\textsubscript{enc} is inspired from multimodal co-attention layer proposed in \cite{yu2019mcan} for visual question answering task.
\begin{equation} \label{SAG-MHAEnc operation}
\begin{split}
 Q^{'} & = LN(Q + \text{Dropout}(MHA(Q,Q,Q))) \\ 
Q^{*} & = MHA_{enc}( Q^{'},K,V) 
\end{split}
\end{equation}
In CM\textsubscript{enc}, the output from the $i$th encoder layer $Enc_{i}$ is passed as query ($Q$) to the subsequent layer with the key ($K$) and value ($V$) remaining the same. Here, $Enc_{i}$ layer can be either MHA\textsubscript{enc} or SAG-MHA\textsubscript{enc}.
\begin{equation} \label{encoder layer operation}
\begin{split}
    Q_{i}&=Enc_{i}(Q_{i-1},K,V) \     i > 0 \\
    Q_{0} &=Enc_{0}(Q,K,V)
\end{split}
\end{equation}
We use separate CM\textsubscript{enc} blocks for processing the foreground and visual scene guided context streams. MCF (MHA\textsubscript{enc}) and MCF (SAG-MHA\textsubscript{enc}) consists of 4 MHA\textsubscript{enc} layers (8 heads and hidden dimension=512) and 3 SAG-MHA\textsubscript{enc} layers (8 heads and hidden dimension=768) in the CM\textsubscript{enc} blocks respectively.
\begin{figure}[h!]
    \centering
    \includegraphics[width=\columnwidth]{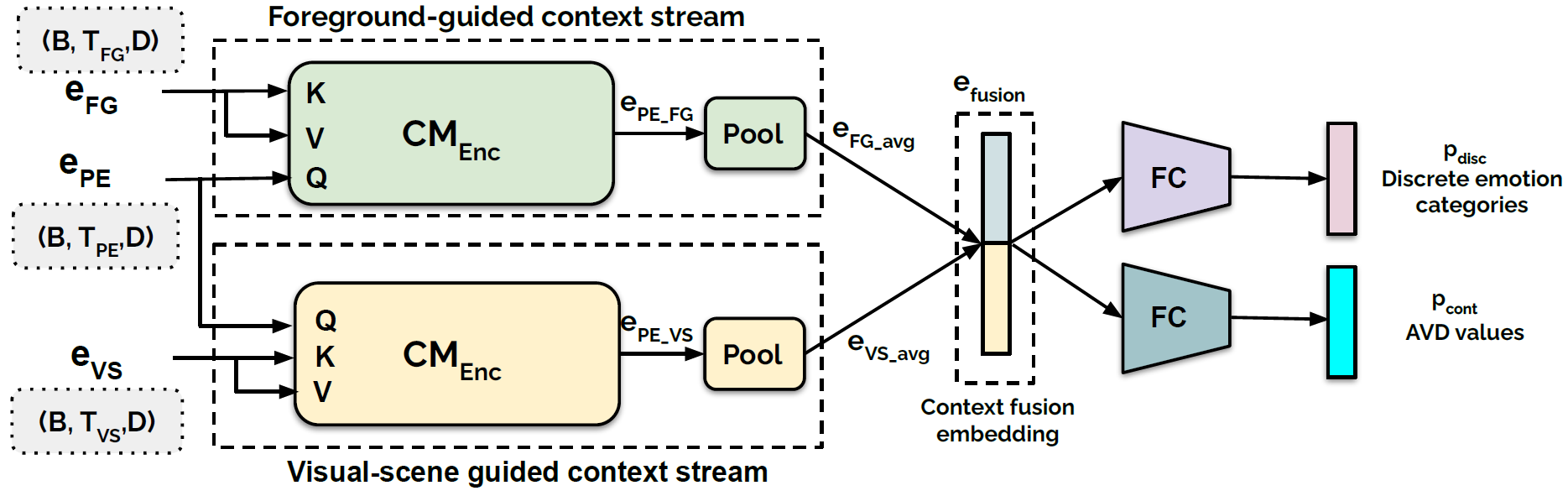}
    \caption{Outline of the $MCF$ module. Separate CM\textsubscript{enc} blocks are used to model the foreground and visual scene context streams. $B$: Batch size. $D$: Hidden dimension. Token length of $T_{FG}$: text representations, $T_{PE}$: person representations, $T_{VS}$: visual scene representations. AVD: Arousal, Valence, Dominance
    }
    \label{mcf module}
\end{figure}
The details of the respective context-guided streams are listed as follows:\\
\textbf{\underline{Foreground-guided context stream:}} We use $e_{PE}$ as query ($Q$) and $e_{FG}$ as key and value inputs to the CM\textsubscript{enc} encoder.\\
\textbf{\underline{Visual-scene guided context stream:}} Similar to the foreground-guided context stream, we use $e_{PE}$ as query ($Q$) and $e_{VS}$ as key and value inputs to the CM\textsubscript{enc} encoder.\\
\textbf{\underline{Context fusion:}}
The outputs from the context-guided streams i.e., $e_{PE\_{FG}}$ and $e_{PE\_{VS}}$ are average pooled and concatenated to obtain a fused embedding as: $e_{fusion}=[e_{FG\_{avg}};e_{VS\_{avg}}]$
\vspace{-3mm}
\section{Experiments}
\textbf{\underline{EMOTIC:}} We use a non-intersecting split of 13584 and 4389 images for training and validation. For testing we use the publicly available split of 5108 images. For joint prediction of 26 discrete emotion classes and the continuous-valued AVD ratings, we use multiple fully-connected (FC) heads in the $MCF$ module with $e_{fusion}$ as input (Fig \ref{mcf module}). The person specific instance in each image is defined by the ground truth person box. We do not consider face as a part of person-specific context since approx 25\% of images do not have visible faces. \\
For training $MCF$ with MHA\textsubscript{enc} and SAG-MHA\textsubscript{enc} layers, we use AdamW \cite{AdamW} (\textit{lr=2e-5}) and Adam \cite{Adam} (\textit{lr=2e-4,  exp($\gamma$=0.90)}) with batch sizes 32 and 64 respectively. \footnote{exp is exponential scheduler \label{note1}}. We use SGD(\textit{lr=1e-2, exp($\gamma$=0.90)}) with a batch size of 64 while training the person-crop only Resnet34 model ($PO_{R34}$) in Table \ref{ablationemotic}. For training all the models associated with EMOTIC, we use a weighted combination of binary-cross entropy (BCE) and mean squared error (MSE) losses.
\begin{equation}
Loss =\lambda_{1} BCE(p_{disc},y_{disc}) + \lambda_{2} MSE(p_{cont},y_{cont})
\end{equation}
Here $y_{disc}$ and $y_{cont}$ refer to ground truth discrete emotion labels and continuous arousal valence dominance ratings. The optimal weights $\lambda_{1}$ and $\lambda_{2}$ are tuned using the validation split. 
\\\\
\textbf{\underline{CAER-S:}} We use a non-intersecting split of 39099 and 9769 video frames across 79 TV shows for training and validation. For testing we use the public split of 20913 video frames. Since face is a dominant signal for persons in TV shows, we use MTCNN \cite{MTCNN} \footnote{https://github.com/timesler/facenet-pytorch} to obtain face crops. We have a single fully-connected (FC) head with $e_{fusion}$ as input for predicting 7 discrete emotion classes. \\
For training $MCF$ with MHA\textsubscript{enc} and SAG-MHA\textsubscript{enc} layers, we use Adam (\textit{lr=2e-4, exp($\gamma$=0.90)}) with a batch size of 64. We use Adam (\textit{lr=1e-4, exp($\gamma$=0.75)}) with a batch size of 64 while training the face-crop only Resnet34 model ($FO_{R34}$) in Table \ref{ablationCAER-S}. For training all the models associated with CAER-S, we use multi-class cross entropy loss.
\\\\
We conduct our experiments using the Pytorch \cite{Paszke2019PyTorchAI} library. We set maximum sequence length $T_{FG}$ as 512 for the captions (Fig \ref{mcf module}). For visual scene and person representations we use $T_{VS}=197$ and $T_{PE}=49$ respectively.
\section{Results}
\subsection{Comparison with state of the art}
We compare performance of $MCF$ (Enc) under two settings where Enc refers to the encoder layer used i.e., MHA\textsubscript{enc} and SAG-MHA\textsubscript{enc} with existing methods in Table \ref{MHAEnc}. We can see that $MCF$ under both settings performs better than prior methods like  \cite{kostiPAMI}, \cite{CAGER}, and \cite{CAER-S} that rely on dual stream (person + whole image approach) and do not use explicit pose information. Furthermore, in contrast to previous methods, we consider language driven foreground (captions) and visual scene contexts instead of end-to-end modeling of whole image based information. For a fair comparison with \cite{Mittal2020EmotiConCM}, the current $MCF$ framework can be potentially expanded to include other person specific streams like face and explicit pose information.
\begin{table}[h!]
\centering
\begin{tabular}{|c|c|}
\hline
\rowcolor[HTML]{DAE8FC} 
\textbf{Model}   & \textbf{mAP}   \\ \hline
Kosti et. al \cite{kostiPAMI} & 27.38          \\ \hline
Zhang et. al  \cite{CAGER}   & 28.42          \\ \hline
Lee et.al  \cite{CAER-S}      & 20.84          \\ \hline
\textbf{MCF (MHA\textsubscript{enc})}     & \textbf{29.53 (0.001)} \\ \hline
MCF (SAG-MHA\textsubscript{enc}) & 28.58 (0.003)  \\ \hline
EmotiCon \cite{Mittal2020EmotiConCM}       & 32.03          \\ \hline
\end{tabular}
\caption{Comparison of $MCF$ module with state of the art in EMOTIC (Test). \textbf{mAP:} mean average precision. Average of runs with 5 random seeds reported with standard deviation for $MCF$.}
\label{MHAEnc}
\end{table}
From Table \ref{MCFSAG}, we can see that in CAER-S, a fully finetuned Resnet34 model trained on face crops ($FO_{R34}$) obtains a high accuracy of 77.35 since facial expressions provide dominant signals for emotion classification in TV shows. However, inclusion of both foreground context through captions and visual scene information in MCF (MHA\textsubscript{enc}) results in better performance (\textbf{79.63}) as compared to both $FO_{R34}$ and baseline attention fusion method CAER-Net-S. 
\begin{table}[h!]
\centering
\begin{tabular}{|c|c|c|}
\hline
\rowcolor[HTML]{DAE8FC} 
\textbf{Model}                               & \textbf{Accuracy}                      & \textbf{F1}                       \\ \hline
CAER-Net-S  \cite{CAER-S}                                & 73.51                             & \textbf{\_}                       \\ \hline
FO\textsubscript{R34}                                    & 77.35 (0.002)                     & 77.13 (0.002)                      \\ \hline
\multicolumn{1}{|l|}{MCF (SAG-MHA\textsubscript{enc})} & \multicolumn{1}{l|}{\textbf{79.63 (0.003)}} & \multicolumn{1}{l|}{\textbf{79.36 (0.003)}}\\ \hline
\end{tabular}
\caption{Comparison of $MCF$ module with state of the art in CAER-S (Test). \textbf{F1:} macro-F1, \textbf{FO:} Face only, \textbf{R34}: Resnet34 fully finetuned. Average of runs with 5 random seeds reported with standard deviation for $MCF$ and Resnet34.}
\label{MCFSAG}
\end{table}
\vspace{-5mm}
\subsection{Ablation studies}
\begin{figure*}
    \centering
    \includegraphics[width=0.80\textwidth]{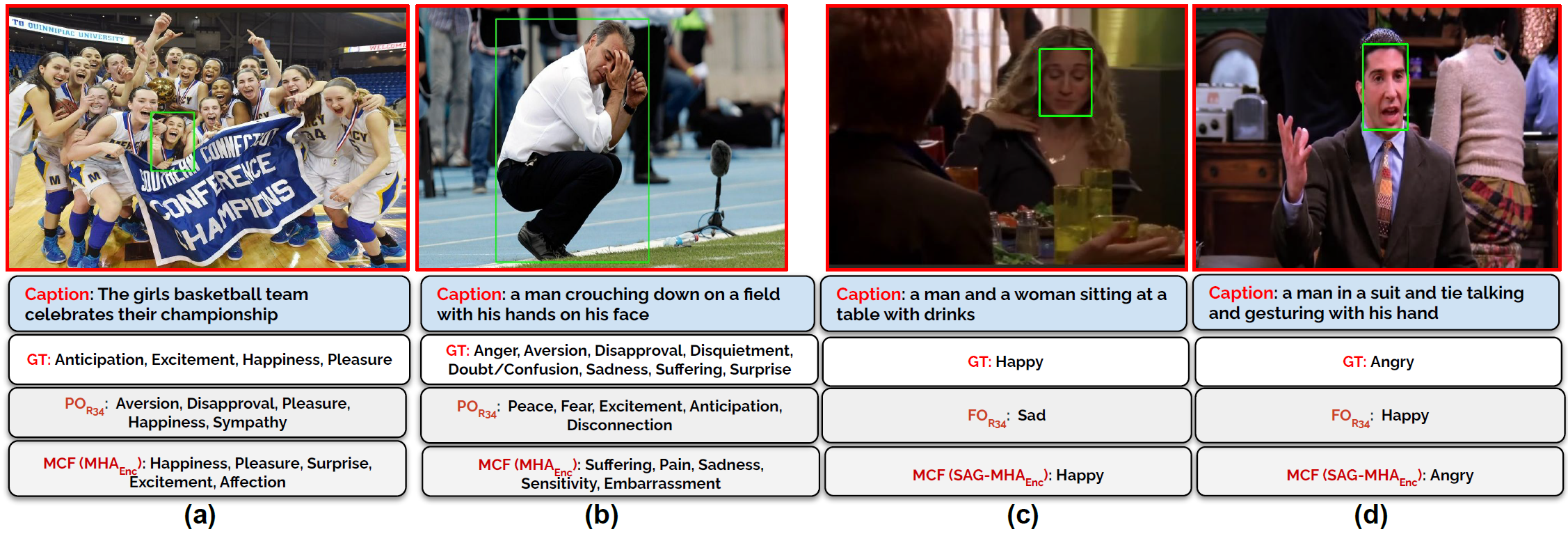}
    \caption{Examples (a) and (b) from EMOTIC showing comparisons between top-5 predictions of $PO_{R34}$ and MCF (MHA\textsubscript{enc}). Examples (c) and (d) from CAER-S showing comparisons between top predictions of $FO_{R34}$ and MCF (SAG-MHA\textsubscript{enc}) . $PO$: Person-only. $FO$: Face-only. $R_{34}$: Resnet34 fully finetuned. GT: Ground truth. Person or face instances marked by green bounding boxes}
    \label{qualfig}
\end{figure*}
We analyze the importance of different input context streams and associated models in EMOTIC. From Table~\ref{ablationemotic}, we can see that the Resnet34 model ($PO_{R34}$) fully finetuned using person specific crops performs worst, thus indicating the need of additional contextual information. Inclusion of scene representation (cls token) from ViT pretrained on Places2 dataset with $PO_{R34}$ via late fusion (LF) improves the mAP to \textbf{25.53}. Freezing $PO_{R34}$ model followed by cross modal interaction through $CM_{enc}$ composed of MHA\textsubscript{enc} layers (visual-scene guided context stream in Fig.~\ref{ablationemotic}) further increases the mAP to \textbf{27.37}. Further we can see that fusion of both foreground and visual scene based context information through MCF (MHA\textsubscript{enc}) results in the best performance (\textbf{29.53}). For both CM\textsubscript{Enc} (MHA\textsubscript{enc} + VS) and MCF (MHA\textsubscript{enc}), we freeze the Resnet34 model for extracting representations from person crops. 
\begin{table}[h!]
\centering
\begin{tabular}{|c|c|c|c|}
\hline
\rowcolor[HTML]{DAE8FC} 
\textbf{Model}                      & \textbf{mAP}            & $\mathbf{\lambda_{1}}$ & $\mathbf{\lambda_{2}}$ \\ \hline
PO\textsubscript{R34}                  & 23.46 (0.006)           & 0.95                   & 0.05               \\ \hline
PO\textsubscript{R34} + VS +LF & 25.53 (0.025)           & 0.6                    & 0.4                \\ \hline
CM\textsubscript{Enc} (MHA\textsubscript{enc} + VS)        & 27.37 (0.004)           & 0.5                    & 0.5                \\ \hline
\textbf{MCF (MHA\textsubscript{enc})}               & \textbf{29.53 (0.001)} & \textbf{0.8}           & \textbf{0.2}       \\ \hline
\end{tabular}
\caption{Ablation study on different context streams and associated models for EMOTIC. \textbf{PO}: Person-only model, \textbf{R34}: Resnet34 fully fine-tuned, \textbf{VS}: Visual scene, \textbf{cls:} cls token, \textbf{LF:} Late fusion, $\mathbf{\lambda_{1}}$: BCE weight, $\mathbf{\lambda_{2}}$: MSE weight. Average of runs with 5 random seeds reported with standard deviation for the models}
\label{ablationemotic}
\end{table}
For CAER-S, we can see from Table \ref{ablationCAER-S} that inclusion of SAG-MHA\textsubscript{enc} layer instead of MHA\textsubscript{enc} improves the accuracy from \textbf{76.89} to \textbf{79.63}.  This can be attributed to the self-attention based augmentation operation for the query features i.e. face representations from Resnet34 in SAG-MHA\textsubscript{enc} layer (Fig \ref{encoder block}). For both MCF (MHA\textsubscript{enc}) and MCF (SAG-MHA\textsubscript{enc}), we finetune the Resnet34 model completely with $MCF$ for extracting representations from face crops. 
\begin{table}[h!]
\centering
\begin{tabular}{|c|c|c|}
\hline
\rowcolor[HTML]{DAE8FC} 
\textbf{Model}  & \textbf{Accuracy} & \textbf{F1}  \\ \hline
FO\textsubscript{R34}          & 77.35 (0.002)     & 77.13 (0.002) \\ \hline
MCF (SAG-MHA\textsubscript{Enc}) & \textbf{79.63 (0.003)}     & \textbf{79.36 (0.003)} \\ \hline
MCF (MHA\textsubscript{Enc})     & 76.89 (0.003)      & 76.74 (0.002) \\ \hline
\end{tabular}
\caption{Ablation study on different context streams and associated models for CAER-S. \textbf{FO}: Face-only model, \textbf{R34}: Resnet-34 fully fine-tuned, Average of runs with 5 random seeds reported with standard deviation for the models}
\label{ablationCAER-S}
\end{table}
\vspace{-5mm}
\subsection{Qualitative examples}
In Fig \ref{qualfig} (a) and (b), we can see that the inclusion of foreground context through captions like \textit{basketball team celebrating} and \textit{man crouching down with his hands on his face} results in consistent performance of MCF (MHA\textsubscript{enc}) as compared to $PO_{R34}$ (Resnet34 finetuned on person crops).  Similarly, for TV shows in Fig \ref{qualfig} (c) while the face crop based prediction from $FO_{R34}$ is \textbf{sad}, inclusion of foreground context with visual scene information gives a correct prediction for MCF (SAG-MHA\textsubscript{enc}). In Fig \ref{qualfig} (d), the act of \textit{gesturing with hand} enables MCF (SAG-MHA\textsubscript{enc}) to make a correct prediction (\textbf{Angry}) as compared to $FO_{R34}$ (Resnet34 finetuned on face crops).
\section{Conclusion}
In this work, we explore the role of contextual information in estimating human emotions with respect to the domains of natural scenes (EMOTIC) and TV shows (CAER-S). Since multimodal-VLN models are pretrained on large-scale image-text pairs from the web, we utilize their capabilities to obtain foreground context information in terms of descriptive captions. Further, we propose a purely attention-based multimodal context fusion (MCF) module to combine person-specific information with the visual scene and foreground context representations. Future work involves the extension of the MCF module to include geometric aspects of person context including pose information and evaluation using media-centered data like movies and advertisements.
\section{Acknowledgement}
We would like to thank the Center for Computational Media Intelligence at USC for supporting this study. 
% References should be produced using the bibtex program from suitable
% BiBTeX files (here: strings, refs, manuals). The IEEEbib.bst bibliography
% style file from IEEE produces unsorted bibliography list.
% -------------------------------------------------------------------------
\bibliographystyle{IEEEbib}
\bibliography{refs}

\begin{thebibliography}{10}

\bibitem{dukes2021}
Daniel Dukes, Kathryn Abrams, Ralph Adolphs, Mohammed~E Ahmed, Andrew Beatty,
  Kent~C Berridge, et~al.,
\newblock ``The rise of affectivism,''
\newblock {\em Nature human behaviour}, vol. 5, no. 7, pp. 816--820, 2021.

\bibitem{AICA}
Sicheng Zhao, Xingxu Yao, Jufeng Yang, Guoli Jia, et~al.,
\newblock ``Affective image content analysis: Two decades review and new
  perspectives,''
\newblock {\em IEEE Transactions on Pattern Analysis and Machine Intelligence},
  vol. 44, no. 10, pp. 6729--6751, 2022.

\bibitem{speechemo}
S.~Latif, R.~Rana, S.~Khalifa, R.~Jurdak, J.~Qadir, and B.~W. Schuller,
\newblock ``Survey of deep representation learning for speech emotion
  recognition,''
\newblock {\em IEEE Transactions on Affective Computing}, , no. 01, pp. 1--1,
  sep 5555.

\bibitem{depressiondetection}
Chiara Zucco, Barbara Calabrese, and Mario Cannataro,
\newblock ``Sentiment analysis and affective computing for depression
  monitoring,''
\newblock in {\em 2017 IEEE International Conference on Bioinformatics and
  Biomedicine (BIBM)}, 2017, pp. 1988--1995.

\bibitem{autismguha}
Tanaya Guha, Zhaojun Yang, Ruth~B. Grossman, and Shrikanth~S. Narayanan,
\newblock ``A computational study of expressive facial dynamics in children
  with autism,''
\newblock {\em IEEE Transactions on Affective Computing}, vol. 9, no. 1, pp.
  14--20, 2018.

\bibitem{savchecnkoengagement}
Andrey~V. Savchenko, Lyudmila~V. Savchenko, and Ilya Makarov,
\newblock ``Classifying emotions and engagement in online learning based on a
  single facial expression recognition neural network,''
\newblock {\em IEEE Transactions on Affective Computing}, pp. 1--12, 2022.

\bibitem{DFEW}
Xingxun Jiang, Yuan Zong, et~al.,
\newblock ``Dfew: A large-scale database for recognizing dynamic facial
  expressions in the wild,''
\newblock in {\em Proceedings of the 28th ACM International Conference on
  Multimedia}, New York, NY, USA, 2020, MM '20, p. 2881–2889, Association for
  Computing Machinery.

\bibitem{Mollahosseini2019AffectNetAD}
Ali Mollahosseini, Behzad Hasani, and Mohammad~H. Mahoor,
\newblock ``Affectnet: A database for facial expression, valence, and arousal
  computing in the wild,''
\newblock {\em IEEE Transactions on Affective Computing}, vol. 10, pp. 18--31,
  2019.

\bibitem{busso2004analysis}
Carlos Busso, Zhigang Deng, Serdar Yildirim, Murtaza Bulut, and Chul Min~others
  Lee,
\newblock ``Analysis of emotion recognition using facial expressions, speech
  and multimodal information,''
\newblock in {\em Proceedings of the 6th international conference on Multimodal
  interfaces}, 2004, pp. 205--211.

\bibitem{BarretEmotionPerception}
Lisa~Feldman Barrett, Batja Mesquita, and Maria Gendron,
\newblock ``Context in emotion perception,''
\newblock {\em Current Directions in Psychological Science}, vol. 20, no. 5,
  pp. 286--290, 2011.

\bibitem{Bar2004VisualOI}
Moshe Bar,
\newblock ``Visual objects in context,''
\newblock {\em Nature Reviews Neuroscience}, vol. 5, pp. 617--629, 2004.

\bibitem{wang2022ofa}
Peng Wang, An~Yang, Rui Men, Junyang Lin, et~al.,
\newblock ``Ofa: Unifying architectures, tasks, and modalities through a simple
  sequence-to-sequence learning framework,''
\newblock {\em CoRR}, vol. abs/2202.03052, 2022.

\bibitem{cho2021vlt5}
Jaemin Cho, Jie Lei, Hao Tan, and Mohit Bansal,
\newblock ``Unifying vision-and-language tasks via text generation,''
\newblock in {\em ICML}, 2021.

\bibitem{kostiPAMI}
Ronak Kosti, Jose~M. Alvarez, Adria Recasens, and Agata Lapedriza,
\newblock ``Context based emotion recognition using emotic dataset,''
\newblock {\em IEEE Transactions on Pattern Analysis and Machine Intelligence},
  vol. 42, no. 11, pp. 2755--2766, 2020.

\bibitem{CAER-S}
Jiyoung Lee, Seungryong Kim, Sunok Kim, Jungin Park, and Kwanghoon Sohn,
\newblock ``Context-aware emotion recognition networks,''
\newblock in {\em 2019 IEEE/CVF International Conference on Computer Vision
  (ICCV)}, 2019, pp. 10142--10151.

\bibitem{Masuda2008PlacingTF}
Takahiko Masuda, Phoebe~C. Ellsworth, Batja Mesquita, et~al.,
\newblock ``Placing the face in context: cultural differences in the perception
  of facial emotion.,''
\newblock {\em Journal of personality and social psychology}, vol. 94 3, pp.
  365--81, 2008.

\bibitem{Dudzik2019ContextIH}
Bernd Dudzik, Michel-Pierre Jansen, Franziska Burger, Frank Kaptein, et~al.,
\newblock ``Context in human emotion perception for automatic affect detection:
  A survey of audiovisual databases,''
\newblock {\em 2019 8th International Conference on Affective Computing and
  Intelligent Interaction (ACII)}, pp. 206--212, 2019.

\bibitem{Wieser2012FacesIC}
Matthias~J. Wieser and Tobias Brosch,
\newblock ``Faces in context: A review and systematization of contextual
  influences on affective face processing,''
\newblock {\em Frontiers in Psychology}, vol. 3, 2012.

\bibitem{AffectNet}
Ali Mollahosseini, Behzad Hasani, and Mohammad~H. Mahoor,
\newblock ``Affectnet: A database for facial expression, valence, and arousal
  computing in the wild,''
\newblock {\em IEEE Transactions on Affective Computing}, vol. 10, no. 1, pp.
  18--31, 2019.

\bibitem{BarsoumICMI2016}
Emad Barsoum, Cha Zhang, Cristian Canton~Ferrer, and Zhengyou Zhang,
\newblock ``Training deep networks for facial expression recognition with
  crowd-sourced label distribution,''
\newblock in {\em ACM International Conference on Multimodal Interaction
  (ICMI)}, 2016.

\bibitem{CAGER}
Minghui Zhang, Yumeng Liang, and Huadong Ma,
\newblock ``Context-aware affective graph reasoning for emotion recognition,''
\newblock in {\em 2019 IEEE International Conference on Multimedia and Expo
  (ICME)}, 2019, pp. 151--156.

\bibitem{Mittal2020EmotiConCM}
Trisha Mittal, Pooja Guhan, et~al.,
\newblock ``Emoticon: Context-aware multimodal emotion recognition using
  frege’s principle,''
\newblock {\em 2020 IEEE/CVF Conference on Computer Vision and Pattern
  Recognition (CVPR)}, pp. 14222--14231, 2020.

\bibitem{pikoulis2021leveraging}
Ioannis Pikoulis, Panagiotis~P. Filntisis, and Petros Maragos,
\newblock ``Leveraging semantic scene characteristics and multi-stream
  convolutional architectures in a contextual approach for video-based visual
  emotion recognition in the wild,''
\newblock in {\em 2021 16th IEEE International Conference on Automatic Face and
  Gesture Recognition (FG 2021)}, 2021.

\bibitem{ALBEF}
Junnan Li, Ramprasaath~R. Selvaraju, et~al.,
\newblock ``Align before fuse: Vision and language representation learning with
  momentum distillation,''
\newblock in {\em NeurIPS}, 2021.

\bibitem{Dosovitskiy2021AnII}
Alexey Dosovitskiy, Lucas Beyer, Alexander Kolesnikov, Dirk Weissenborn,
  et~al.,
\newblock ``An image is worth 16x16 words: Transformers for image recognition
  at scale,''
\newblock {\em ArXiv}, vol. abs/2010.11929, 2021.

\bibitem{zhou2017places}
Bolei Zhou, Agata Lapedriza, Aditya Khosla, Aude Oliva, and Antonio Torralba,
\newblock ``Places: A 10 million image database for scene recognition,''
\newblock {\em IEEE Transactions on Pattern Analysis and Machine Intelligence},
  2017.

\bibitem{He2016DeepRL}
Kaiming He, X.~Zhang, Shaoqing Ren, and Jian Sun,
\newblock ``Deep residual learning for image recognition,''
\newblock {\em 2016 IEEE Conference on Computer Vision and Pattern Recognition
  (CVPR)}, pp. 770--778, 2016.

\bibitem{Devlin2019BERT}
Jacob Devlin, Ming-Wei Chang, Kenton Lee, and Kristina Toutanova,
\newblock ``Bert: Pre-training of deep bidirectional transformers for language
  understanding,''
\newblock in {\em NAACL}, 2019.

\bibitem{wolf-etal-2020-transformers}
Thomas Wolf, Lysandre Debut, Victor Sanh, Julien Chaumond, et~al.,
\newblock ``Transformers: State-of-the-art natural language processing,''
\newblock in {\em Proceedings of the 2020 Conference on Empirical Methods in
  Natural Language Processing: System Demonstrations}, Online, Oct. 2020, pp.
  38--45, Association for Computational Linguistics.

\bibitem{yu2019mcan}
Zhou Yu, Jun Yu, Yuhao Cui, Dacheng Tao, and Qi~Tian,
\newblock ``Deep modular co-attention networks for visual question answering,''
\newblock in {\em Proceedings of the IEEE Conference on Computer Vision and
  Pattern Recognition (CVPR)}, 2019, pp. 6281--6290.

\bibitem{AdamW}
Ilya Loshchilov and Frank Hutter,
\newblock ``Decoupled weight decay regularization,''
\newblock in {\em ICLR}, 2019.

\bibitem{Adam}
Diederik~P. Kingma and Jimmy Ba,
\newblock ``Adam: A method for stochastic optimization,''
\newblock {\em CoRR}, vol. abs/1412.6980, 2015.

\bibitem{MTCNN}
Kaipeng Zhang, Zhanpeng Zhang, Zhifeng Li, and Yu~Qiao,
\newblock ``Joint face detection and alignment using multitask cascaded
  convolutional networks,''
\newblock {\em IEEE Signal Processing Letters}, vol. 23, pp. 1499--1503, 2016.

\bibitem{Paszke2019PyTorchAI}
Adam Paszke, Sam Gross, Francisco Massa, et~al.,
\newblock ``Pytorch: An imperative style, high-performance deep learning
  library,''
\newblock {\em ArXiv}, vol. abs/1912.01703, 2019.

\end{thebibliography}

\end{document}